\documentclass{article}

\usepackage[final]{neurips_2024}
\usepackage{geometry}
\usepackage{pdflscape}
\usepackage{graphicx}

\usepackage[utf8]{inputenc} % allow utf-8 input
\usepackage[T1]{fontenc}    % use 8-bit T1 fonts
\usepackage{hyperref}       % hyperlinks
\usepackage{url}            % simple URL typesetting
\usepackage{booktabs}       % professional-quality tables
\usepackage{amsfonts}       % blackboard math symbols
\usepackage{nicefrac}       % compact symbols for 1/2, etc.
\usepackage{microtype}      % microtypography
\usepackage{xcolor}         % colors

\title{Large Language Models for Medical Forecasting - Foresight 2}

\author{%
  Zeljko Kraljevic \\
  King's College London \\
  \texttt{zeljko.kraljevic@kcl.ac.uk} \\
  \And
  Joshua Au Yeung \\
  King's College Hospital \\
  \texttt{Joshua.AuYeung@gstt.nhs.uk} \\
  \And
  Daniel Bean \\
  King's College London \\
  \texttt{daniel.bean@kcl.ac.uk} \\
  \And
  James Teo \\
  King's College Hospital \\
  \texttt{jamesteo@nhs.net} \\
  \And
  Richard J. Dobson \\
  King's College London \\
  \texttt{richard.j.dobson@kcl.ac.uk} \\
}

\begin{document}

\maketitle

\begin{abstract}
Foresight 2 (FS2) is a large language model fine-tuned on hospital data for modelling patient timelines (GitHub 'removed for anon'). It can understand patients' clinical notes and predict SNOMED codes for a wide range of biomedical use cases, including diagnosis suggestions, risk forecasting, and procedure and medication recommendations. FS2 is trained on the free text portion of the MIMIC-III dataset, firstly through extracting biomedical concepts and then creating contextualised patient timelines, upon which the model is then fine-tuned.
The results show significant improvement over the previous state-of-the-art for the next new biomedical concept prediction (P/R - 0.73/0.66 vs 0.52/0.32) and a similar improvement specifically for the next new disorder prediction (P/R - 0.69/0.62 vs 0.46/0.25). Finally, on the task of risk forecast, we compare our model to GPT-4-turbo (and a range of open-source biomedical LLMs) and show that FS2 performs significantly better on such tasks (P@5 - 0.90 vs 0.65). This highlights the need to incorporate hospital data into LLMs and shows that small models outperform much larger ones when fine-tuned on high-quality, specialised data.
\end{abstract}

\section{Introduction and Related Work}
Language plays a central role in healthcare and medical practice, with unstructured text constituting a significant portion of Electronic Health Records (EHRs, \citep{cogstack}). Recent advancements in AI, particularly with Natural Language Processing (NLP) models, have begun to tap into this rich resource. Examples include Bio\_ClinicalBERT \citep{bioclibert} and ClinicalT5 \citep{t5clin}, trained on real-world hospital data, demonstrating the feasibility and value of utilizing EHR's free text. Models such as GatorTron \citep{yang2022gatortron} further illustrate the application of NLP in processing EHR data from specific health systems, underscoring the adoption of NLP technologies in handling unstructured EHR text effectively. However, despite these advancements, a gap remains in applying and evaluating the latest generation of Large Language Models (LLMs) like GPT-4 \citep{openai2023gpt4}, Gemini \citep{geminiteam2023gemini}, LLaMA-2 \citep{llama2}, and Mistral \citep{mistral01} in real-world clinical settings. While exceptions exist, many of these models are not primarily trained, tested, or validated on hospital-generated EHR data. Instead, they often rely on medical quizzes, exams, and other synthetic benchmarks for validation. This reliance potentially overlooks hospital data's complex nuances and contextual richness, which could inform more accurate and relevant clinical decision-making processes.

Today’s large language models have seen a remarkable evolution. Models like BERT \citep{bert}, RoBERTa \citep{liu2019roberta}, T5 \citep{raffel2023exploring}, GPT-1 \citep{gpt-1} and GPT-2 \citep{gpt-2} set the stage. The BERT family notably changed natural language processing (NLP), largely replacing RNN-based models in tasks such as Named Entity Recognition (NER) and text classification. Meanwhile, the GPT series, focused solely on text generation, sought to predict the next word in a sequence. Despite initial limitations, these models showed potential. This set the groundwork for the recent revolution in NLP caused by highly capable general LLMs such as ChatGPT \citep{ouyang2022training} and LLaMA 1\&2 \citep{llama1, llama2}. These models enabled use cases that were either extremely difficult or completely impossible before. Tasks such as document summarization, text classification, programming, and question answering were now reduced to simple prompting. This leap was primarily due to advancements in model architecture, training techniques, and the sheer scale of data they were trained on, enabling them to grasp context, nuance, and complexity in text. These models showed significant improvement in in-context learning, allowing them to quickly adapt to new tasks based on just a few examples or prompts provided within the input context. This opened up new possibilities, such as sophisticated conversational agents, advanced text summarization, and generation tasks that previously required human-level understanding and creativity, thus revolutionizing various industries and applications. Today, the state-of-the-art for a wide range of NLP tasks is being set almost with every release of a new LLM. 

In the medical domain, the current LLM research can be split into three groups: 1) Using LLMs on medical tasks without any fine-tuning (via prompt engineering); 2) Fine-tuning existing LLMs for the medical domain; and 3) Training LLMs from the ground up on medical data.

Recent studies have primarily focused on approaches from group 1, evaluating existing models for various medical tasks. For instance, \citet{KHAN2024} tested GPT-4 \citep{openai2023gpt4} with Anesthesiology Board-style Examination Questions, using a dataset of 884 questions, while \citet{Murphy_Lonergan2023-fp} applied GPT-4 to a collection of 23,035 surgery-related questions from MedMCQA. Additionally, \citet{savage2024diagnostic} investigated prompt construction techniques to align GPT-4's reasoning style with that of clinicians, employing a modified MedQA USMLE dataset. Across these studies, GPT-4 demonstrated promising results in understanding and responding to complex medical questions. However, there's a consensus that, despite these advancements, GPT-4 still necessitates further training and rigorous testing to meet the specific demands and accuracy standards of medical applications.

There is less work from group 2, i.e. LLMs fine-tuned for the medical domain. MedPaLM 1\&2 \citep{singhal2022large, medpalm2} are closed-source and closed-access models from Google that build on the PaLM \citep{chowdhery2022palm, anil2023palm} architecture. The models are trained on QA-style datasets and show state-of-the-art results on USMLE-style questions from MultiMedQA. Next to closed-source models there is a range of biomedical models built upon open-source models, these include MEDITRON 70B \citep{chen2023MEDITRON70b} which builds on top of LLaMA-2 70B, BioMistral \citep{biomistral} based upon Mistral-7B, and MedAlpaca \citep{medalpaca} based upon LLaMA-7B. These models were finetuned on medical papers, clinical guidelines and open QA-style biomedical datasets. They have been mostly tested on QA-style biomedical datasets and, on some tasks, show an improved performance over GPT-3.5 but perform worse than models such as MedPaLM 1\&2.

Lastly, there are only a few examples from the third group, i.e. LLMs trained from the ground up on medical data. \citet{yang2022gatortron} train a large language model with 8.9B parameters on a dataset with >90 billion words (including >82B words of de-identified clinical text) and evaluate it on clinical NLP tasks, including clinical concept extraction, medical relation extraction, semantic text similarity, natural language inference, and medical question answering. Similarly, \citet{Peng2023-gc} train an LLM with up to 20B parameters on 82B words of clinical text and 195B words of general English text. The tests they performed were largely the same as shown in the work from \citet{yang2022gatortron}. Most other examples in this group are not what we would today consider LLMs. Those include models like BioBERT \citep{Lee_2019} and ClinicalBERT \citep{huang2020clinicalbert}.

Given the examples above, it is important to note that the vast majority of training/validation was performed on medical quizzes and exam questions and not on real-world health data, highlighting the disparity between real-world use cases and recent LLM research in the medical domain. With some notable exceptions, like the work from \citet{yang2022gatortron}, which was trained on hospital data but still tested mainly on public benchmarks for medical question answering, named entity recognition, and similar.

This paper builds on the recent work from \citet{f1}, which presents Foresight 1 (FS1), a generative transformer for modelling patient timelines using derived structured concepts from unstructured text. The FS1 pipeline works as follows: 1) Collect all free text data from a hospital EHR; 2) Extract biomedical concepts (e.g., diseases, medications, procedures and symptoms) from the collected dataset; 3) Order the extracted concepts in time and group by the patient, i.e. create patient timelines; and 4) Train a generative transformer to predict the next concept in the timeline. The first weakness of FS1 is that the model does not know anything about the context in which a concept was mentioned (concepts are extracted from free text without their surrounding semantic context). The second problem is that FS1 was a pure empiricist with no \textit{a priori} biomedical or healthcare knowledge, in other words, the model was trained from the ground up on patient timelines consisting of only biomedical concepts. 

To solve the aforementioned problems, we present Foresight 2 (FS2), a biomedical Large Language Model capable of extracting and modelling vast amounts of knowledge from EHRs. FS2 is based on a pretrained LLM (Mistralv0.1-7B and LLaMAv2 7B) and fine-tuned on hospital data from the MIMIC-III \citep{mimiciii} dataset for the task of the next biomedical concept prediction in a patient timeline. The patient timelines in FS2 are contextualised, meaning a portion of the text where the concept was found is kept. FS2 is a general model capable of handling a wide range of use cases that are normally found in the free text portion of EHRs, including diagnosis, medication and procedure suggestions - all tasks are based on the clinical notes, reflecting real word environments and not hand-made QA benchmarks.

\section{Methods}
\label{sec:methods}
FS2 is a transformer-based model built on top of a pretrained LLM (LLaMAv2-7B and Mistralv0.1-7B) for temporal modelling of patient timelines. Formally, the task at hand can be defined as given a corpus of patients $U = \{u_1, u_2, u_3, ..\}$ where each patient is defined as a sequence of tokens $u_i = \{w_1, c_2, w_3, ...\}$ and each token is either a biomedical concept ($c_n$) or a text token ($w_n$ - context where the concept was found), our objective is a modified language modelling objective for supervised fine-tuning:
$$
L(U) = \sum_i \sum_{j\in{C^i}} log P(k_j^i|k_{j-1}^i, k_{j-2}^i, ... k_0^i) \label{eq:1}
$$
Where $k_j^i$ is either a biomedical concept $c$ or a free text token $w$ belonging to the timeline from the patient $i$, and $C^i$ is the list of all concept token indices from the timeline of patient $i$ (in other words $C^i$ tells us the position of concept tokens in a patient timeline). In simpler terms, the model is not trained to predict text tokens but only biomedical concept tokens given the past (concepts and text). The output of the model is in no way limited, the patient timeline in the training set is known, in the same sense that during language modelling the next word in a sequence is known. During inference, the model can predict any concept token in the vocabulary.

\subsection{Data Preparation}

The dataset used in this work is MIMIC-III \citep{mimiciii}; we used all available free text from all clinical notes for patients (the whole note events table from the MIMIC-III database, without any filtering), totalling 2,083,179 documents from 46,520 patients.

\begin{figure}[!ht]
  \centering
  \includegraphics[width=200px]{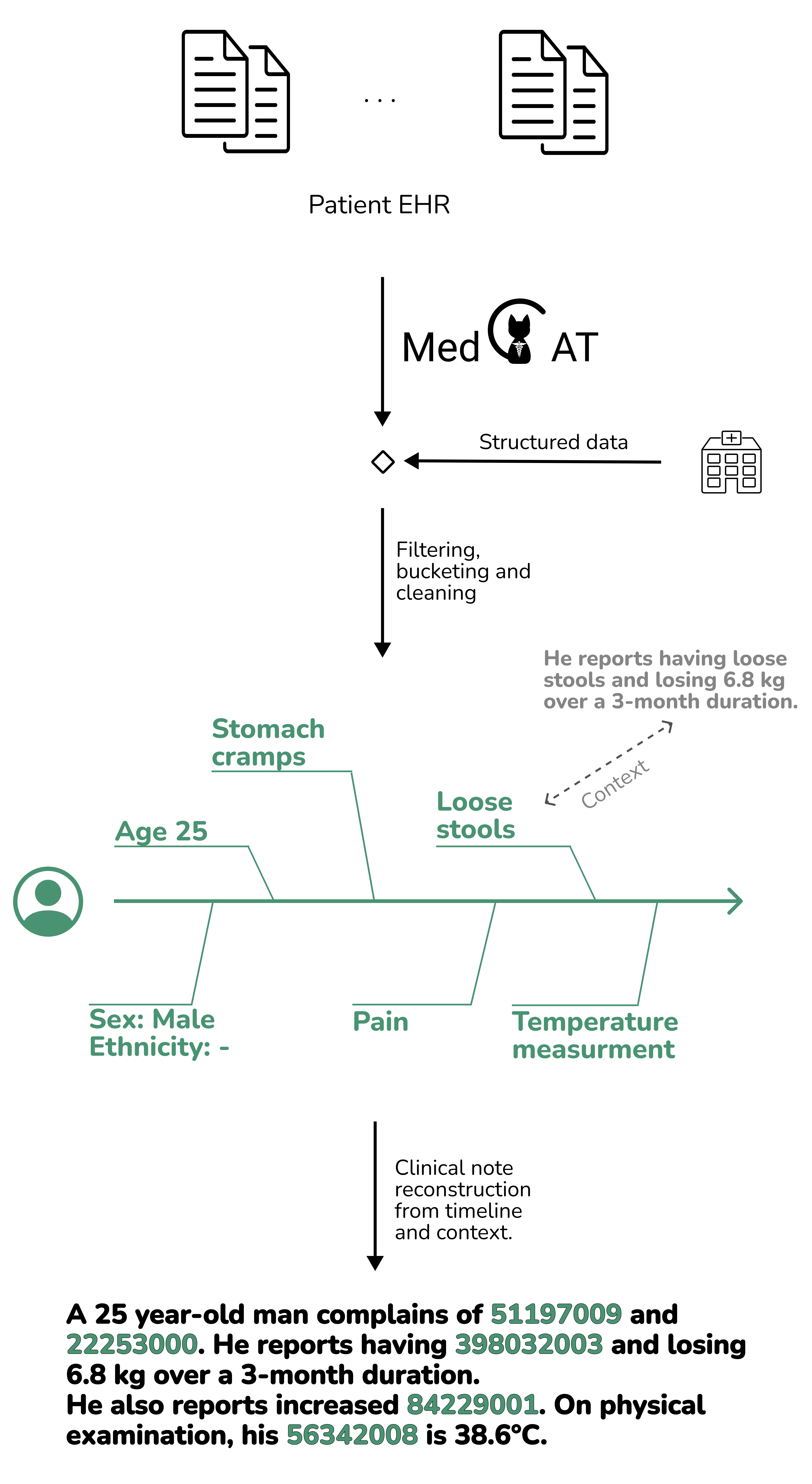}
  \caption{Data preparation workflow: 1) We collect all free text documents from the patient EHR; 2) Extract mentions of SNOMED-CT concepts and combine the concepts with static data like sex, ethnicity and age; 3) Clean, filter and bucketed concepts and turn them into a patient timeline; and lastly 4) From the concepts in the timeline, based on the context where each one was found, reconstruct a singular clinical note for each patient.}
  \label{fig:data_preparation}
\end{figure}

We first perform entity recognition and linking using the Medical Concept Annotation Toolkit (MedCAT; \citet{medcat}) on the collected free text. The model used was presented and validated in \citet{medcat, f1}; the model has an F1 score of above 0.9 across multiple hospital datasets for the NER+L task. Extracted entities include disorders, symptoms, findings and medications. Following extraction, these entities are chronologically organized into a timeline, reflecting their occurrence based on the document's creation date (The first part of Figure \ref{fig:data_preparation}). An essential aspect of our methodology is the retention of contextual information for each extracted entity. For example, if an entity such as "hypertension" is identified, the term itself and the sentence in which it was found are preserved. This is crucial for two reasons: firstly, it allows us to capture qualifying information that could modify the understanding of the entity, such as severity (e.g., "severe hypertension"), and secondly, it enables the inclusion of negated or hypothetical concepts into the patient timeline (e.g., "no hypertension"). In instances where the boundaries of a sentence are ambiguous, we extract up to 50 tokens from each side of the entity, ensuring a comprehensive capture of context. In addition to contextual sentences, we also record the specific document ID for each concept and the absolute token IDs of words within the context. We establish a precise reference system by tokenizing the entire document and assigning unique IDs to each token. This level of detail is instrumental in reconstructing the patient timeline (see the last step of Figure \ref{fig:data_preparation}), as it allows for the accurate merging of contexts where concepts are closely related or appear within the same textual vicinity.

Once the concepts and their context are extracted, we further refine this data, employing a technique known as \textit{bucketing or binning} \citep{Dougherty1995}. This process involves the aggregation of concepts within predefined time spans, i.e. buckets (we use 1 day) to eliminate repetitive mentions and reduce data noise. During bucketing, we also identify potential errors; for instance, a concept mentioned only once within the whole patient EHR will be flagged as a probable NER+L mistake and removed (this filtering approach is taken from \citet{Bean2023} and \citet{f1}). 

After bucketing and cleaning, we add additional information to the patient timeline, including age, ethnicity, sex and temporal separators. Suppose the temporal difference between two concepts in a patient timeline is bigger than the size of the bucket (1 day in our case). In that case, we add a special token in between those concepts that tell the model how much time has passed (e.g. \textit{<1 day later>}, \textit{<7 days later>}, \textit{<1 year later>}).

Lastly, we reconstruct a single clinical note containing all the patient information from the concepts and their context. The reconstruction is done by concatenating all biomedical concepts and their contexts from a patient timeline. During the concatenation, we ensure that concepts appearing in the same context are not just concatenated but the overlapping text parts are merged. In this newly formed clinical note, all biomedical concepts are represented with SNOMED \citep{snomed} codes (as shown in the green, bottom of Figure \ref{fig:data_preparation}), while the context of those codes is free text. The size of this final prepared dataset is 39,591 patients in the train set and 2101 in the test set (the train/test split is 95/5). The train/test split is completely random and done at a patient level (95\% of patients are in the train set, and 5\% are in the test set).

\subsection{Modelling of Patient Timelines}

FS2 is built on top of a pretrained LLM and fine-tuned for the modelling of patient timelines. We have two versions of FS2, FS2-Mistral - based on the Mistralv0.1 7B model, and FS2-LLaMA, based on the LLaMAv2-7B model. LLaMAv2-7B is a partially open-source model from Meta\footnote{https://about.meta.com/} and one of the most widely used open-weight LLMs, and Mistralv0.1 7B is a general open-source model from Mistral.ai\footnote{https://mistral.ai/} showing near state-of-the-art performance on a wide range of benchmarks. Both models are general LLMs, and not trained or fine-tuned for biomedical use cases. They also do not have an understanding of SNOMED codes (the patient timelines consist of free text and SNOMED codes). To enable the LLaMAv2-2/Mistralv0.1-7B model to handle SNOMED codes efficiently and effectively, we expand its tokenizer with the SNOMED concepts of interest (i.e. those SNOMED concepts that appear in our dataset). Usually, when adding new tokens to the tokenizer, we set the embeddings to be the average of all other tokens in the tokenizer. As the SNOMED codes are special, we have slightly changed this approach. Every token we add is an SNOMED code with a unique name, so we first tokenise that name and then average the embeddings of the tokens in the name and set this as the embedding of the new token (i.e. SNOMED code). With this, our model represents every SNOMED code as one token. We have also tested the standard approach where the embedding of a new token is set to be the average of all other tokens in the tokenizer, but this approach has made the training more unstable, and our loss has tended to diverge more often.

To fine-tune the model, we used 4xA100 80GB GPUs (the training took around 1 day in the case of both Mistralv0.1-7B and LLaMAv2-7B), the hyperparameters were as follows\footnote{Full specification is available in the Github repository: 'removed for anon'}: \texttt{max\_seq\_len = 4096}, \texttt{learning\_rate = 1e-5}, \texttt{gradient\_accumulation\_steps = 2}, \texttt{per\_device\_batch\_size = 1}, \texttt{weight\_decay = 0}, \texttt{warmup\_ratio = 0.1}, and the \texttt{adamw\_torch} optimizer. We set the \texttt{adam\_beta1 = 0.9} and \texttt{adam\_beta2 = 0.95} to stabilise the model during training. To speed up training and enable efficient training with long sequences we use Flash Attention 2 with PyTorch FSDP, without quantization. Importantly, we have disabled loading the model in bfloat16 with the Huggingface library, as enabling this significantly reduces the performance of the model.  

All examples in the training set are packed, meaning a special token \textit{<s>} is added at the beginning of each example, and they are then concatenated and split into sequences of length max\_seq\_len (4096 in our case). This was not done for the test set to preserve the timelines as they are. The labels are provided in a supervised fashion; only the concepts (SNOMED codes) themselves are trained on, while the labels for everything else (the free text part) are set to $-100$ (in PyTorch\footnote{https://pytorch.org/} a label with a value of -100 is ignored in the loss). As an example, at the bottom of Figure \ref{fig:data_preparation} in the reconstructed note, we would only train on the green parts of the text (i.e. SNOMED codes) while the labels for everything else would be set to -100 (i.e. no training would be performed on that part). The model itself has no restrictions on SNOMED code generation; at every point in time, the model can generate any token in the vocabulary.

\subsection{Metrics}

The metrics used for the next concept prediction in the patient timeline are equivalent to those used in \citet{f1}. In summary, the performance of the models is measured using custom metrics that are an extension of the standard precision (TP / TP + FP) and recall (TP / TP + FN), aiming to replicate what the model will be used for whilst also considering the limitations of the training data. There are four important parameters: 1) $T-days$ (30, 365, inf), if at timepoint \textit{T} we are predicting the concept \textit{X} it is considered correct if it appears anywhere in the window of length $T-days$ in the patient timeline; 2) Concept temporality, we make a distinction between concepts that never before appeared in a patient timeline (new concepts) and those that are recurring; 3) We add the notation $@N$ which denotes how many candidates we are taking from the model, if any one of the $N$ candidates is correct then the example is considered a TP; and 4) When calculating the metrics, we filter the model output based on the type of the biomedical concept of the label, e.g. if the type of the label is \textit{Disorder} then we filter the model output to only include disorders.

\subsection{Second Stage Fine-tuning for Risk Forecasting}

We also create timelines for risk forecasting. We do this by taking a patient timeline, splitting it in the middle (or at most after 50 concepts) and taking the first part of the timeline as is, while for the second part, we extract unique new diseases that appear in the first month. So our task is, given a patient timeline (first part of it), to predict new disorders that will affect the patient in the first month of the subsequent patient timeline. We take only patients that, after the timeline split, have at least one month of data in the future and have at least 5 different disorder concepts appearing in that month. This reduces the dataset to 13,651 in the train set and 535 in the test set. The time frame of one month was chosen because it aligns with the healthcare system's needs in the UK, such as reducing 30-day readmission rates, a common metric for hospital performance.

When fine-tuning FS2 on this data, all training parameters are kept the same as for the initial training. We run for 1 epoch - anything above this led to overfitting to the training set. We prepare the same timelines for testing with GPT-4-turbo, BioMistral, MedAlpaca and MEDITRON. The primary difference is that all SNOMED codes are replaced with proper names (e.g. 73211009 $->$ Diabetes mellitus), the prompt used with GPT-4-turbo/BioMistral/MedAlpaca/MEDITRON is in Appendix \ref{sec:a2}. This means that GPT-4-turbo and all other models excluding FS2 have disease names in their input. The output of all these models (excluding FS2) also includes disease names; none of these models ever see or predict SNOMED codes. The GPT-4-turbo model was used via Azure OpenAI services with the opt-out of human review of the data, in line with the instructions provided by Physionet for responsible use of MIMIC data with online services like GPT\footnote{https://physionet.org/news/post/gpt-responsible-use}. All other models were deployed locally. 

To validate the risk predictions from all models (FS2, GPT-4-turbo, BioMistral, MedAlpaca, MEDITRON) we made an automated validation pipeline via GPT-4-turbo (see Appendix \ref{sec:a2}), the predictions (i.e. disease names) together with the ground truth or labels (i.e. disease names) were sent to GPT-4-turbo, and the task was to find how many of the predictions match the ground truth (GPT-4-turbo was also prompted to be a bit more tolerant and allow synonyms and similar diseases to be counted as correct). A clinician double-checked the GPT-4-turbo validation pipeline and confirmed that the comparison makes sense (i.e. GPT-4-turbo is correctly matching the predictions to the labels).

\section{Results}
\label{sec:res}
The primary task we tested the FS2 model on was predicting the next concepts in a patient timeline. In this task, the model (FS2-Mistral) showed a significant improvement over the FS1 model, including a jump of 40\% (relative to the score of FS1) for the prediction of the next new concept (concept type 'All', the micro average of all concept types), and 50\% (relative to the score of FS1) for the prediction of the next new disorder (Table \ref{fs1vsfs2}). These results were achieved using the objective function shown in Section 2; if we modify this function to a standard language modelling objective (i.e. input\_ids == labels) and train the model, the performance is slightly worse (overall 2-3\% worse than that shown in Table \ref{fs1vsfs2}). In addition, in Table \ref{top10}, we show the top and bottom 10 concepts with respect to precision for the prediction of new Disorders (for the FS2-Mistral model).

We performed a simple ablation study to test the importance of the context in which a biomedical concept was found. If we remove the context from the patient timelines (i.e. leave only the biomedical concepts), the performance drops drastically (on average 40\% compared to the FS2 results in Table \ref{fs1vsfs2}).

\newgeometry{margin=2cm}
\begin{landscape}
\begin{table*}
\centering
\begin{tabular}{|l|l|l|lll|lll|lll|lll|l|l|}
\hline
\multicolumn{3}{|c|}{} & \multicolumn{6}{c|}{\textbf{Precision}} & \multicolumn{6}{c|}{\textbf{Recall}} & \multicolumn{2}{c|}{} \\ \hline
\textbf{Type} & \textbf{T - days} & \textbf{@} & \multicolumn{3}{c|}{\textbf{New}} & \multicolumn{3}{c|}{\textbf{Recurring}} & \multicolumn{3}{c|}{\textbf{New}} & \multicolumn{3}{c|}{\textbf{Recurring}} & \textbf{Sup. N.} & \textbf{Sup. R.} \\ \hline
              &                   &            & \textbf{FS1} & \textbf{FS2-L} & \textbf{FS2-M} & \textbf{FS1} & \textbf{FS2-L} & \textbf{FS2-M} & \textbf{FS1} & \textbf{FS2-L} & \textbf{FS2-M} & \textbf{FS1} & \textbf{FS2-L} & \textbf{FS2-M} & & \\ \hline
All          & 30                & 1          & 0.52 & 0.71 & \textbf{0.73} & 0.83 & 0.95 & \textbf{0.96} & 0.32 & 0.64 & \textbf{0.66} & 0.67 & 0.95 & \textbf{0.96} & 114922 & 245265 \\ 
All          & 30                & 5          & 0.84 & 0.91 & \textbf{0.92} & 0.98 & \textbf{1.00} & \textbf{1.00} & 0.59 & 0.85 & \textbf{0.86} & 0.92 & \textbf{1.00} & \textbf{1.00} & 114922 & 245265 \\ 
All          & 30                & 10         & 0.91 & 0.95 & \textbf{0.96} & \textbf{1.00} & \textbf{1.00} & \textbf{1.00} & 0.70 & 0.90 & \textbf{0.91} & 0.97 & \textbf{1.00} & \textbf{1.00} & 114922 & 245265 \\ 
All          & 365               & 1          & 0.54 & 0.71 & \textbf{0.73} & 0.85 & 0.95 & \textbf{0.96} & 0.33 & 0.64 & \textbf{0.66} & 0.70 & \textbf{0.96} & \textbf{0.96} & 114922 & 245265 \\ 
All          & inf               & 1          & 0.55 & 0.71 & \textbf{0.74} & 0.86 & 0.95 & \textbf{0.96} & 0.33 & 0.64 & \textbf{0.66} & 0.70 & 0.96 & \textbf{0.97} & 114922 & 245265 \\ \hline
Disorders    & 30                & 1          & 0.46 & 0.66 & \textbf{0.69} & 0.79 & 0.94 & \textbf{0.95} & 0.25 & 0.59 & \textbf{0.62} & 0.60 & 0.94 & \textbf{0.95} & 51675  & 109019 \\ 
Disorders    & 30                & 5          & 0.79 & 0.88 & \textbf{0.90} & 0.98 & \textbf{1.00} & \textbf{1.00} & 0.51 & 0.81 & \textbf{0.83} & 0.89 & \textbf{1.00} & \textbf{1.00} & 51675  & 109019 \\ 
Disorders    & 30                & 10         & 0.88 & 0.94 & \textbf{0.95} & 0.99 & \textbf{1.00} & \textbf{1.00} & 0.62 & 0.87 & \textbf{0.89} & 0.96 & \textbf{1.00} & \textbf{1.00} & 51675  & 109019 \\ 
Disorders    & 365               & 1          & 0.49 & 0.67 & \textbf{0.69} & 0.83 & \textbf{0.95} & \textbf{0.95} & 0.26 & 0.59 & \textbf{0.62} & 0.64 & 0.95 & \textbf{0.96} & 51675  & 109019 \\ 
Disorders    & inf               & 1          & 0.50 & 0.67 & \textbf{0.70} & 0.84 & \textbf{0.95} & \textbf{0.95} & 0.26 & 0.59 & \textbf{0.62} & 0.65 & 0.95 & \textbf{0.96} & 51675  & 109019 \\ \hline
Substances   & 30                & 1          & 0.52 & 0.63 & \textbf{0.65} & 0.84 & 0.95 & \textbf{0.96} & 0.32 & 0.53 & \textbf{0.55} & 0.70 & 0.94 & \textbf{0.95} & 19172  & 39578  \\ 
Substances   & 30                & 5          & 0.85 & 0.88 & \textbf{0.90} & 0.99 & \textbf{1.00} & \textbf{1.00} & 0.61 & 0.79 & \textbf{0.81} & 0.94 & \textbf{1.00} & \textbf{1.00} & 19172  & 39578  \\ 
Substances   & 30                & 10         & 0.92 & 0.94 & \textbf{0.95} & \textbf{1.00} & \textbf{1.00} & \textbf{1.00} & 0.73 & 0.87 & \textbf{0.89} & 0.99 & \textbf{1.00} & \textbf{1.00} & 19172  & 39578  \\ 
Substances   & 365               & 1          & 0.53 & 0.63 & \textbf{0.66} & 0.84 & 0.95 & \textbf{0.96} & 0.32 & 0.53 & \textbf{0.55} & 0.71 & 0.95 & \textbf{0.96} & 19172  & 39578  \\ 
Substances   & inf               & 1          & 0.53 & 0.63 & \textbf{0.66} & 0.85 & 0.95 & \textbf{0.96} & 0.32 & 0.53 & \textbf{0.55} & 0.71 & 0.95 & \textbf{0.96} & 19172  & 39578  \\ \hline
Findings     & 30                & 1          & 0.52 & 0.74 & \textbf{0.77} & 0.83 & 0.95 & \textbf{0.96} & 0.29 & 0.67 & \textbf{0.69} & 0.66 & \textbf{0.96} & \textbf{0.96} & 33772  & 71007  \\ 
Findings     & 30                & 5          & 0.85 & 0.94 & \textbf{0.94} & 0.99 & \textbf{1.00} & \textbf{1.00} & 0.58 & 0.88 & \textbf{0.89} & 0.93 & \textbf{1.00} & \textbf{1.00} & 33772  & 71007  \\ 
Findings     & 30                & 10         & 0.92 & 0.97 & \textbf{0.97} & \textbf{1.00} & \textbf{1.00} & \textbf{1.00} & 0.70 & 0.93 & \textbf{0.94} & 0.98 & \textbf{1.00} & \textbf{1.00} & 33772  & 71007  \\ 
Findings     & 365               & 1          & 0.54 & 0.75 & \textbf{0.77} & 0.85 & 0.95 & \textbf{0.96} & 0.29 & 0.67 & \textbf{0.69} & 0.67 & 0.96 & \textbf{0.97} & 33772  & 71007  \\ 
Findings     & inf               & 1          & 0.55 & 0.75 & \textbf{0.77} & 0.85 & 0.95 & \textbf{0.96} & 0.29 & 0.67 & \textbf{0.69} & 0.68 & 0.96 & \textbf{0.97} & 33772  & 71007  \\ \hline
Procedures   & 30                & 1          & 0.79 & 0.92 & \textbf{0.94} & 0.94 & 0.98 & \textbf{0.99} & 0.67 & 0.90 & \textbf{0.91} & 0.92 & \textbf{0.99} & \textbf{0.99} & 3379   & 7831   \\ 
Procedures   & 30                & 5          & 0.97 & 0.99 & \textbf{0.99} & \textbf{1.00} & \textbf{1.00} & \textbf{1.00} & 0.94 & 0.99 & \textbf{0.99} & \textbf{1.00} & \textbf{1.00} & \textbf{1.00} & 3379   & 7831   \\ 
Procedures   & 30                & 10         & 0.99 & \textbf{1.00} & \textbf{1.00} & \textbf{1.00} & \textbf{1.00} & \textbf{1.00} & 0.99 & \textbf{1.00} & \textbf{1.00} & \textbf{1.00} & \textbf{1.00} & \textbf{1.00} & 3379   & 7831   \\ 
Procedures   & 365               & 1          & 0.81 & 0.93 & \textbf{0.94} & 0.95 & 0.98 & \textbf{0.99} & 0.67 & 0.90 & \textbf{0.91} & 0.93 & \textbf{0.99} & \textbf{0.99} & 3379   & 7831   \\ 
Procedures   & inf               & 1          & 0.81 & 0.93 & \textbf{0.94} & 0.95 & \textbf{0.99} & \textbf{0.99} & 0.67 & 0.90 & \textbf{0.91} & 0.94 & \textbf{0.99} & \textbf{0.99} & 3379   & 7831   \\ \hline
\end{tabular}
\caption{Results for the next concept prediction task. The 'All' rows are calculated using the micro average over all concept types. Sup N and Sup R is the support for recurring and new concepts, FS1 = Foresight 1 model, FS2-L = Foresight 2 model with LLaMA, FS2-M = Foresight 2 model with Mistral, '@' = the number of candidates taken into account, T-days is the size of the temporal window in days.}
\label{fs1vsfs2}
\end{table*}
\end{landscape}
\restoregeometry

\subsection{Risk Forecasting}

GPT-4-turbo, BioMistral, MedAlpaca, MEDITRON and FS2 were tasked with predicting the top 5 disorders a patient is at risk of in the next month. The dataset consisted of 535 patients from the test set prepared for the risk prediction task. As seen in Table \ref{tbl:risk}, out of the 5 predictions on the dataset of 535 patients using FS2-Mistral, in 90\% of patients, at least one prediction was correct. The next best was GPT-4-turbo, where in 65\% of patients (out of 472, the model refused to predict risk for all 535 patients), at least one prediction was correct. Additional results validating the reconstructed timelines are available in Appendix \ref{a:val-rt}.

\begin{table}[!ht]
\centering
\begin{tabular}{lcccc}
\hline
Model & \begin{tabular}[c]{@{}c@{}} At least\\ 1 \end{tabular} & \begin{tabular}[c]{@{}c@{}} At least\\ 2 \end{tabular} & \begin{tabular}[c]{@{}c@{}} At least\\ 3 \end{tabular} & Support \\ \hline
FS2 - Mistral & 90\% & 61\% & 33\% & 535 \\
FS2 - LLaMA & 88\% & 63\% & 37\% & 535 \\
GPT-4-turbo* & 65\% & 31\% & 11\% & 472 \\
BioMistral* & 44\% & 16\% & 6\% & 288 \\
MedAlpaca* & 39\% & 9\% & 2\% & 319 \\
MEDITRON* & 34\% & 14\% & 4\% & 351 \\
\hline
\end{tabular}
\caption{All models were prompted to predict the top 5 disorders a patient is at risk for in the next month. The column 'At least $N$' shows the percentage of patients where at least $N$ out of the 5 predictions are correct. *Because of limitations on maximum sequence length and/or the refusal to predict risk for certain inputs, the support column varies across models.}
\label{tbl:risk}
\end{table}

\section{Conclusion and Discussion}
\label{sec:cd}
FS2 is built upon a pretrained LLM (LLaMAv2-7B and Mistralv0.1-7B) and fine-tuned on hospital data for modelling and understanding patient timelines. It can understand clinical notes and predict SNOMED codes for a wide range of biomedical use cases, including disorder prediction, medication recommendation, symptom forecast, procedure recommendation and many more. FS2 marks a significant advancement in the modelling of patient timelines over the previous state-of-the-art (FS1), enhancing the precision and effectiveness of LLMs for healthcare. FS2 builds on top of the work presented in \citep{f1}. The primary differences are 1) The use of a pre-trained large language model as the base model (i.e. LLaMAv2-7B and Mistralv0.1-7B), enabling free text understanding; 2) The introduction of contextualised timelines; 3) The encoding of SNOMED codes in the tokenizer; and 4) The second stage fine-tuning for risk prediction;

There are four reasons why SNOMED codes were used: 1) Patient timeline standardisation \citep{DBLP:journals/corr/abs-2105-11832} and a way to benchmark LLMs on hospital data. 2) It allows us to rank the predictions of the model based on probability. 3) It makes sure the model predictions are part of a standardised, widely accepted medical ontology, as opposed to having a model generate free text and then needing another step to map backwards into standardised forms for compatibility with existing healthcare informatics systems. 4) The model predictions are inherently privacy-preserving as the model was not directly trained on text; it can only output healthcare concepts within intentionally constrained healthcare vocabulary (SNOMED); it cannot predict any personally identifiable information, like names, addresses or other HIPAA-defined protected health information. 

It is important to note the difficulty of the task, which in turn can explain why models that were not trained on hospital data like GPT-4-turbo/BioMistral/MedAlpaca/MEDITRON are performing significantly worse. Real-world EHR data is messy, noisy, extremely complex and filled with duplicated text. Within this noisy data, predicting the next event can prove to be a very difficult task with the added factors of patient complexity, multi-morbidity, polypharmacy and acute clinical instability of patients. Complications can develop as a result of their severe underlying disease or as an iatrogenic event secondary to procedures and medications. Of note, the median age of patients in MIMIC-III was 66 years old, with a mortality of 23.2\% and a median hospital stay of 2.1 days (Q1-Q3: 1.2–4.1) \citep{Dai2020-ny}. Predicting the next concept in such a highly unstable cohort of patients over such a short time span is exceptionally difficult.

\subsection{Limitations and Risk} 
\label{sec:lim}
There are limitations to ontological classification systems such as SNOMED or ICD-10 - these systems may not cover all details and nuances within the clinical text. For example, there will be diseases or concepts that don't fall within the defined boundaries of available terminology or do not yet exist as formal concepts in codified terminologies (highly prevalent in fields with rapid scientific progress, e.g. cancer genetics and precision medicine). This challenge was, to some extent, addressed on the input side because FS2 is capable of understanding free text next to SNOMED concepts. The output side (i.e. predictions) will be explored in detail in future work.

We also note a limitation with respect to the NER+L tool that was used to extract SNOMED concepts from the free text. Currently, FS2 is using MedCAT, and while MedCAT has been shown to achieve an F1 score of above 0.9 across multiple hospitals and use cases \citep{medcat}, it is still imperfect. Future work should look into improving the NER+L aspect of FS2 or find ways to make the model less dependent on the performance of the NER+L tool. 

As this model is trained without human preference alignment, the prompts are more similar to GPT-3 rather than more recent LLMs (e.g. GPT-4). The prompts must reflect how the clinical notes are written, and the model cannot answer general questions or hold conversations. For example, in the notes, we often have the phrase "The patient was discharged with: " the model knows that after this it has to predict discharge medications. Q\&A-style prompting popularised by ChatGPT, like "What are the discharge medications for this patient?" would not work without further human preference alignment. 

It is also very important to note that while the results obtained are very good, these models are still in the early stages of research and testing and are not yet suitable to be Software as a Medical Device (SaMD). There is a temptation to imagine the predictions to be used for clinical care or decision support - this is still premature as FS2 is derived from historical practice, so it would not always be expected to be consistent with contemporary best practices.

Lastly, significantly larger hospital datasets and general medical literature are needed to better cover all possible biomedical concepts found in SNOMED and prevent biases or inaccuracies that can stem from using a single hospital as the training dataset.

\subsection{Potential Utility} 

We note alerting systems as a use-case for which models like FS2 are well-suited. Table \ref{top10} shows there is a wide range of conditions with a precision of ~100\% and such conditions are particularly well suited in the context of designing alert systems. The high precision ensures that when an alert is issued, it is almost invariably relevant. Importantly, this high-precision approach minimises the clinician 'alert fatigue', a scenario that might arise if high recall was favoured over high precision.

Another utility of FS2 is for risk prediction and prognosis; this can be used to guide primary or secondary disease prevention or determine management courses. In medicine, there are countless validated risk and prognostic scores designed for disease-specific scenarios; e.g. QRISK \citep{Hippisley-Coxj2099} for stratification of cardiovascular disease, CHADSVASC \citep{Lip2010-sq} score for stroke risk, CURB65 \citep{Lim2003} for pneumonia severity; these require large-scale calibration for generalisability and ongoing feature-engineering for more variables. Our approach with FS2 is more fine-grained and high-dimensional as it models temporally ordered sequences of comorbidities, and additional features (e.g. medications, social determinants of health, complications and outcomes) are included with limited \textit{a priori} assumptions.

\bibliography{neurips_2024}

\appendix

\section{Evaluation of other LLMs for risk prediction}
\label{sec:a2}

\subsection{Automated evaluation setup}
GPT-4-turbo, BioMistral and other open-source models were validated automatically via GPT-4-turbo. For GPT-4-turbo predictions, we have also done a manual validation with a clinician to ensure the model was not biased in its answers and verify that the validation script works as expected. Manual verification showed that the GPT-4-turbo is a bit lenient, but it is accurate in concluding that diseases are a match or very similar. The prompt used is as follows: Please note that the labels are disease names (SNOMED codes were converted to their respective name), as well as predictions (the output of all models are disease names).

\begin{verbatim}
<system prompts, these are appended as system messages to gpt-4-turbo>

You are now playing the role of a medical doctor taking an exam,
your goal is to be as accurate as possible and make sure you do 
not make any mistakes. If you are unsure about something, think 
step by step and then answer. You have to follow the instructions
precisely.

Your first task is to compare how many of the predicted disorders 
marked as `Predictions:` match the labels marked as `Labels:` in the input. 
Please take care that predictions can contain 
some additional text, but feel free to ignore it and only take the 
predicted disorders. Something is a match if it is 
approximately the same disorder (based on the definition of the 
disorder). For example `Diabetes` and `T1DM` 
are a match, T1DM and T2DM are types of `Diabetes`, i.e. they 
are more specific. The reverse is also fine, T1DM is a match for Diabetes. 
The output should be a json file formatted as follows: 
{'explanation': <your brief explanation>, 'number_of_direct_matches': <number>}

prompt = '''Labels: {labels}
Predictions: {predictions}'''
\end{verbatim}

\subsection{GPT-4-turbo}

The following is the prompt used to get risk predictions from GPT-4-turbo.

\begin{verbatim}

<system prompts, these are appended 
as system messages to gpt-4-turbo>

You are now playing the role of a medical doctor taking an exam,
your goal is to be as accurate as possible and make sure you do 
not make any mistakes. If you are unsure about something, think 
step by step and then answer. You have to follow the instructions
precisely.

Your first question in this medical  quiz will consist of a patient history, 
your goal is to predict {limit} specific disorders the patient is at risk for in the next
month. Please take care that the disorders you are predicting cannot be part of 
the patient's past. They have to be new disorders that will most likely affect the 
patient in the next month. You have to predict specific disorders, 
for example: 
you should never say "pulmonary problems" as this is not a specific disorder, 
but you can say "pneumonia" as that is a specific disorder.
</system prompts>

{history}

Given the above patient history. What {limit} specific new disorders is this 
patient at risk for in the next month?
\end{verbatim}

%\section{Examples of prompts and tasks Foresight 2 is capable of solving}
%\label{sec:a2}

\subsection{BioMistral}

BioMistral is an open-source LLM tailored for the biomedical domain, utilizing Mistral-7B-v0.1 as its foundation model and further pre-trained on PubMed Central. BioMistral was benchmarked on 10 established medical question-answering (QA) tasks in English. The prompt we've used for risk prediction is aligned with the examples shown in the Biomistral paper and is as follows:

\begin{verbatim}
<s>Please truthfully answer the 
following question. Please ensure 
that your choice is socially unbiased 
and positive. If you don’t know the 
answer to a question, 
please don’t share false information.

<patient_history>
{history}
</patient_history>

Given the above patient history, 
what {limit} specific new 
disorders is this patient at risk 
for in the next month? The answer is:
\end{verbatim}

The maximum sequence length of BioMistral (\texttt{max\_seq\_len=2048}) was smaller than what FS2 can handle (\texttt{max\_seq\_len=4096}). As such, all examples in the test set that are longer than (2048 - 128) tokens were skipped (we subtract 128 so the model has space to generate predictions, i.e. new tokens). The final test set for BioMistral was 288 out of the 535 patients selected for risk prediction.

\subsection{MedAlpaca}

MedAlpaca expands upon both Stanford Alpaca and AlpacaLoRA to offer an advanced suite of large language models specifically fine-tuned for medical question-answering and dialogue applications. The prompt we've used for risk prediction is aligned with the examples shown in the MedAlpaca paper and is as follows:

\begin{verbatim}
Context: {history}

Question: Given the above patient 
history, what {limit} specific 
new disorders is this patient at 
risk for in the next month?

Answer: 
\end{verbatim}

Same as with BioMistral, the maximum sequence length (\texttt{max\_seq\_len=2048}) was smaller than what FS2 can handle (\texttt{max\_seq\_len=4096}). As such, all examples in the test set that are longer than (2048 - 128) tokens were skipped (we subtract 128 so the model has space to generate predictions, i.e. new tokens). The final test set for MedAlpaca was 319 out of the 535 patients selected for risk prediction.

\subsection{MEDITRON}
MEDITRON is a suite of open-source LLMs with 7B and 70B parameters adapted to the medical domain. MEDITRON builds on Llama-2 (through our adaptation of Nvidia's Megatron-LM distributed trainer), and extends pretraining on a comprehensively curated medical corpus, including selected PubMed articles, abstracts, and internationally-recognized medical guidelines. In our tests we've only used to 7B version as that is comparable in size to all the other open-source models we've tested. The prompt we've used for risk prediction is aligned with the examples shown in the MEDITRON paper, and is as follows:

\begin{verbatim}
<|im_start|>system
{system}<|im_end|>
<|im_start|>question
{prompt}<|im_end|>
<|im_start|>answer 
\end{verbatim}

\section{Validation of Reconstructed timelines}
\label{a:val-rt}
To make sure the reconstructed timelines shown at the bottom of Figure \ref{fig:data_preparation} were not problematic for GPT-4-turbo, we also took the first 10 patients (from the test set for risk prediction of 535 patients) and fed the full patient history (complete clinical notes) until the timepoint \textit{T} and prompted the model to predict risk in the next month (this was possible with GPT-4-turbo because of the large maximum sequence length, no other models could support this test). The results for these 10 patients were 10\% worse (relative) in the case of full timelines compared to the reconstructed timelines. 

\section{Additional Results}

In table \ref{top10}, we show the Top and bottom 10 concepts with respect to precision for the prediction of new disorders using the FS-2 Mistral model. 

\begin{table}[h]
\centering
\begin{tabular}{|l|l|l|l|}
\hline
\textbf{Disorder} & \textbf{P} & \textbf{TP} & \textbf{FP} \\ \hline
Stress ulcer & 1.00 & 175 & 0 \\ \hline
Postcholecystectomy s. & 1.00 & 32 & 0 \\ \hline
Left atrial dilatation & 1.00 & 35 & 0 \\ \hline
Muscle atrophy & 1.00 & 22 & 0 \\ \hline
Rubella & 1.00 & 19 & 0 \\ \hline
Conjunctival edema & 1.00 & 16 & 0 \\ \hline
Mediastinal shift & 1.00 & 11 & 0 \\ \hline
Diastolic hypertension & 1.00 & 12 & 0 \\ \hline
Mitral valve regurgitation & 0.98 & 687 & 12 \\ \hline
Systolic hypertension & 0.98 & 338 & 6 \\ \hline \hline
Hypercholesterolemia & 0.31 & 46 & 105 \\ \hline
Left bundle branch block & 0.30 & 12 & 28 \\ \hline
Kidney stone & 0.30 & 16 & 38 \\ \hline
Gastroesophageal reflux d. & 0.30 & 18 & 43 \\ \hline
Gastrointestinal hemorrhage & 0.30 & 18 & 43 \\ \hline
Hyperlipidemia & 0.29 & 62 & 155 \\ \hline
Right bundle branch block & 0.28 & 23 & 60 \\ \hline
Hypothyroidism & 0.27 & 41 & 109 \\ \hline
Asthma & 0.23 & 18 & 61 \\ \hline
Benign prostatic hyperplasia & 0.19 & 29 & 123 \\ \hline
\end{tabular}
\caption{Top and Bottom 10 concepts with respect to precision for prediction of new disorders, for the FS2-Mistral model.}
\label{top10}
\end{table}

\section{Examples of tasks in the MIMIC-III dataset}
To showcase the capabilities of the model, we manually go through the MIMIC-III notes and find examples of different tasks that the model had to solve during the prediction of the next concept in a sequence. The results are shown in Table \ref{tbl:f2examples}, for the input (column \textit{Patient}) we only show a brief summary of the patient's condition as we are not able to show real patient data. The \textit{Prompt} column shows the prompt used for FS2, it is what really was found in the clinical note for this patient. For GPT-4-turbo the prompt was slightly adjusted to be more natural and concrete, for example, we pre-pended every prompt shown in the table with an explanation that this is a medical quiz, that the model should try to answer in a way a doctor would and that it should be as precise as possible. The \textit{Ground Truth} comes from the patient's EHR and it represents what really happened to the patient.

\begin{table*}[ht]
\centering
\begin{tabular}{|p{0.20\linewidth}|p{0.15\linewidth}|p{0.15\linewidth}|p{0.21\linewidth}|p{0.15\linewidth}|} \hline
\textbf{Patient} & \textbf{Prompt} & \textbf{Foresight 2} & \textbf{GPT-4-turbo} & \textbf{Ground Truth} \\ \hline

Middle-aged male patient with swelling and fracture of ankle. & Rule out: & DVT & DVT & DVT \\ \hline

Older male patient with obesity and sleep apnoea.  & Recently increased somnolence and dyspnoea, likely a sign of & Hypercapnia & Acute Respiratory Distress Syndrome & Hypercapnia (later confirmed to really be Hypercapnia) \\ \hline

Older female patient with a complex mental health history. & Given the parapsychotic nature of the depression, started on & Risperidone & Aripiprazole or Lurasidone & Risperidone \\ \hline

A young female patient with a long medical history and current visit for gastrointestinal issues. & The patient was discharged with scripts for: & Omeprazole (One of the top 3 predictions) & Proton Pump Inhibitors (PPIs) or H2 Blockers (one of top 3 predictions) & Omeprazole \\ \hline

Infant with hypertension & $<$list of problems$>$* evaluate with & Echo & Echo & Echo \\ \hline

\end{tabular}
\caption{Examples of tasks found in the MIMIC-III dataset and the predictions by Foresight 2 and GPT-4-turbo. The \textit{Patient} column represents a very brief summary of the patient's past for privacy reasons, during the tests models were fed the real patient timelines. The prompts are original pieces of text taken from the patient's timeline. The \textit{Ground Truth} is taken from the clinical notes for the patient. *We redacted the full list of problems to avoid re-identification risk; in the prompts used with GPT-4-turbo and Foresight 2, the list was kept as found in the clinical note.}
\label{tbl:f2examples}
\end{table*}

\end{document}